\documentclass[journal]{IEEEtran}
\usepackage{amsmath,amsfonts}
\usepackage{algorithmic}
\usepackage{algorithm}
\usepackage{array}
\usepackage[caption=false,font=normalsize,labelfont=sf,textfont=sf]{subfig}
\usepackage{booktabs}
\usepackage{textcomp}
\usepackage{stfloats}
\usepackage{url}
\usepackage{verbatim}
\usepackage{graphicx}
\usepackage{cite}
\usepackage{color}
\usepackage{soul} %
\usepackage{color, xcolor} %
\usepackage{pifont}
\begin{document}

\title{Knowledge Guided Entity-aware Video Captioning and A Basketball Benchmark}
\author{Zeyu Xi, Ge Shi,  Xuefen Li, Junchi Yan,~\IEEEmembership{Senior Member,~IEEE}, Zun Li,  Lifang Wu$^{\ast}$\thanks{*Corresponding author},~\IEEEmembership{Senior Member,~IEEE}, Zilin Liu, Liang Wang,~\IEEEmembership{Fellow,~IEEE}
\thanks{This work was supported in part by the Natural Science Foundation of China under Grant 62106010, Grant 62236010, Grant 62306022; in part by the Beijing Natural Science Foundation under grant L233008; in part by the China Postdoctoral Science Foundation under Grant 2022M720318; Beijing Postdoctoral Science Foundation under Grant 2022-zz-077.2.

Zeyu Xi, Ge Shi, Xuefen Li, Zun Li, Lifang Wu and Zilin Liu are with Faculty of Information Technology, Beijing University of Technology, Beijing 100124, China (e-mail: 
xi961226@163.com; shige@bjut.edu.cn; lixuefen@emails.bjut.edu.cn; zunli@bjut.edu.cn; lfwu@bjut.edu.cn; zl\_irene0113@163.com)

Junchi Yan is with Department of Computer Science and Engineering, and MoE Key Lab of Artificial Intelligence, AI Institute, Shanghai Jiao Tong University, Shanghai 200240, China (e-mail: yanjunchi@sjtu.edu.cn)

Liang Wang is with Chinese Academy of Sciences (CASIA), Beijing, China (e-mail: liang.wang@ia.ac.cn)
}
}
\maketitle

\begin{abstract}
Despite the recent emergence of video captioning models, how to generate the text description with specific entity names and fine-grained actions is far from being solved, which however has great applications such as basketball live text broadcast. In this paper, a new multimodal knowledge graph supported basketball benchmark for video captioning is proposed. Specifically, we construct a multimodal basketball game knowledge graph (KG\_NBA\_2022) to provide additional knowledge beyond videos. Then, a multimodal basketball game video captioning (VC\_NBA\_2022) dataset that contains 9 types of fine-grained shooting events and 286 players' knowledge (i.e., images and names) is constructed based on KG\_NBA\_2022. We develop a knowledge guided entity-aware video captioning network (KEANet) based on a candidate player list in encoder-decoder form for basketball live text broadcast. The temporal contextual information in video is encoded by introducing the bi-directional GRU (Bi-GRU) module. And the entity-aware module is designed to model the relationships among the players and highlight the key players. Extensive experiments on multiple sports benchmarks demonstrate that KEANet effectively leverages extera knowledge and outperforms advanced video captioning models. The proposed dataset and corresponding codes will be publicly available soon. 
\end{abstract}


\section{Introduction}
\IEEEPARstart{V}{ideo} captioning (VC) is a crucial computer vision task that requires the model to outputs corresponding text descriptions based on the given video. By associating visual information and text elements, this task has been blossoming and gaining increasing attention owing to its many promising applications, such as video title automatic generation \cite{2020Comprehensive}, visually-impaired assistance \cite{huo2021wenlan}, video storytelling \cite{Storytelling} and online video search \cite{DBLP:conf/mm/NieQM00B22, DBLP:journals/corr/abs-2309-11091}.

Despite the recent rapid development of video captioning \cite{mahmud2023clip4videocap, lin2022swinbert,luo2020univl, liu2021o2na, STAT, Show, yang2022clip}, these methods are not practical in real-world scenarios. They cannot automatically generate a text description with specific entity names and fine-grained actions. As the basketball live text broadcast example in Fig. 1, the basketball video involves multi-person actions and complex scenes, which pose significant challenges to model's performance and generalization. Note that conventional models can only generate the simple sentence to describe the video from a macroscopic perspective (e.g., a man fails to make a shot and another man gets the rebound). In contrast, if the model has game-related knowledge such as players who appear in the game and fine-grained actions in basketball, it can generate the knowledge-grounded text description (e.g., Brandon Ingram misses the 2pt jump shot and Justise Winslow gets the defensive rebound). In addition, existing common used benchmarks, including MSVD \cite{chen2011collecting}, YouCook \cite{das2013thousand}, MSR-VTT~\cite{xu2016msr}, and ActivityNet Captions~\cite{krishna2017dense}, simplify the task by using indefinite pronouns like ``a man", ``a woman" or ``a group of men" instead of specific entity names. And the actions in the annotated captions are coarse-grained. These benchmarks cannot provide relevant knowledge beyond videos and fine-grained action annotations to develop models to generate text descriptions with specific entity names and fine-grained actions. 
\begin{figure}[tb!]
	\centering
	\includegraphics[width=8.8cm,height=4.1cm]{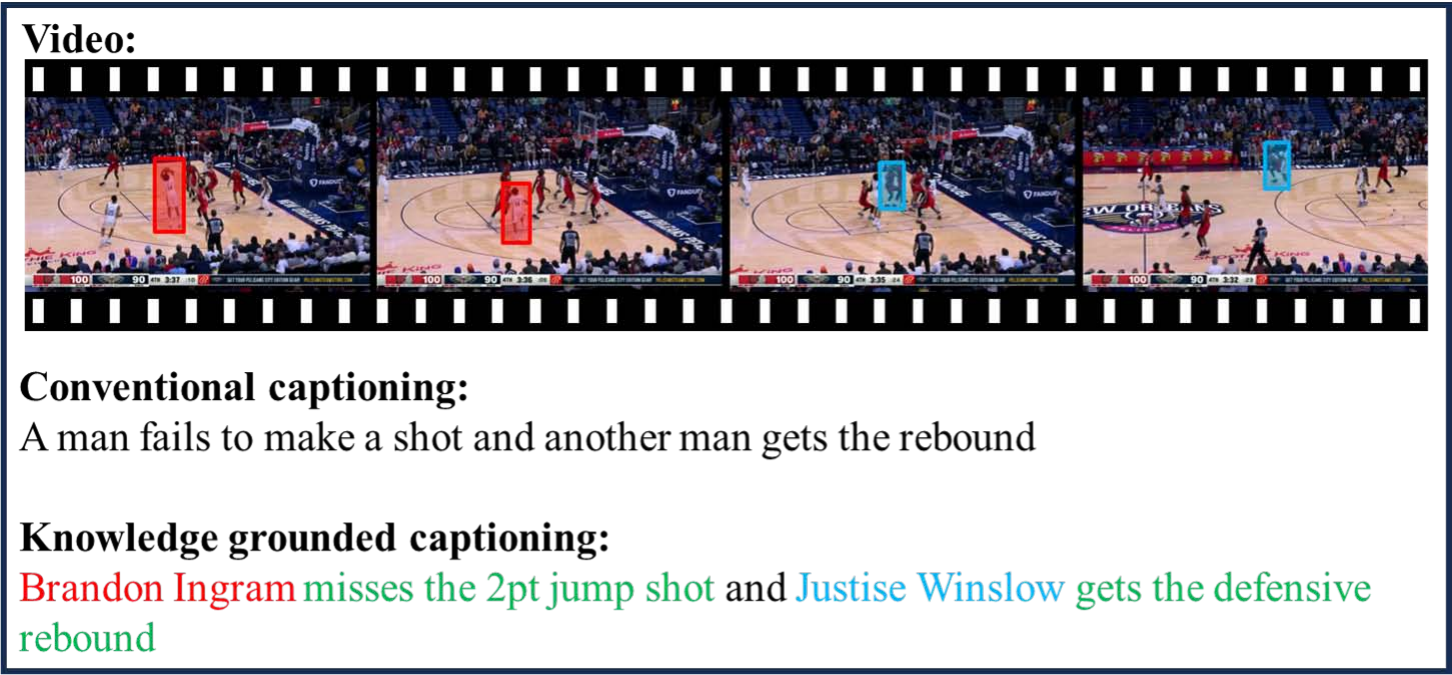}
	\caption{Comparison of conventional captioning with knowledge-grounded captioning. The different specific entity names are marked {\color{red}red} and {\color{blue}blue}, respectively. And the fine-grained actions are marked {\color{green}green}.}
	\label{fig:Figure 1}
\end{figure}

An increasing number of researchers have discovered that conventional methods and benchmarks for video captioning fail to meet the requirements of practical applications and have made numerous attempts to address this issue. Mkhallati et al. \cite{mkhallati2023soccernet} publicly release SoccerNet-Caption, the first dataset for dense video captioning in soccer broadcast videos. Although this work focuses on researching entity-aware video captioning, it cannot provide any additional knowledge for the model to generate text descriptions with specific entity names. Ayyubi et al. \cite{ayyubi2023video} first train an entity perception detector to detect entities in video frames. They then utilize a large language model to retrieve relevant knowledge and enrich the visual content, thereby generating entity-aware text descriptions for news videos summary. 
However, this approach is limited by the performance of entity perception detector and large language model. Furthermore, Qi et al. \cite{qi2023goal} build a unimodal knowledge graph and a benchmark (Goal) for soccer commentary. The provided knowledge is not comprehensive and lacks visual aspects. It can be seen that these methods are tailored to different domains. The introduction of relevant background knowledge is essential to achieve text descriptions with specific entity names in a particular field.

\begin{figure*}[tb!]
	\centering
	\includegraphics[width=17.7cm,height=9.5cm]{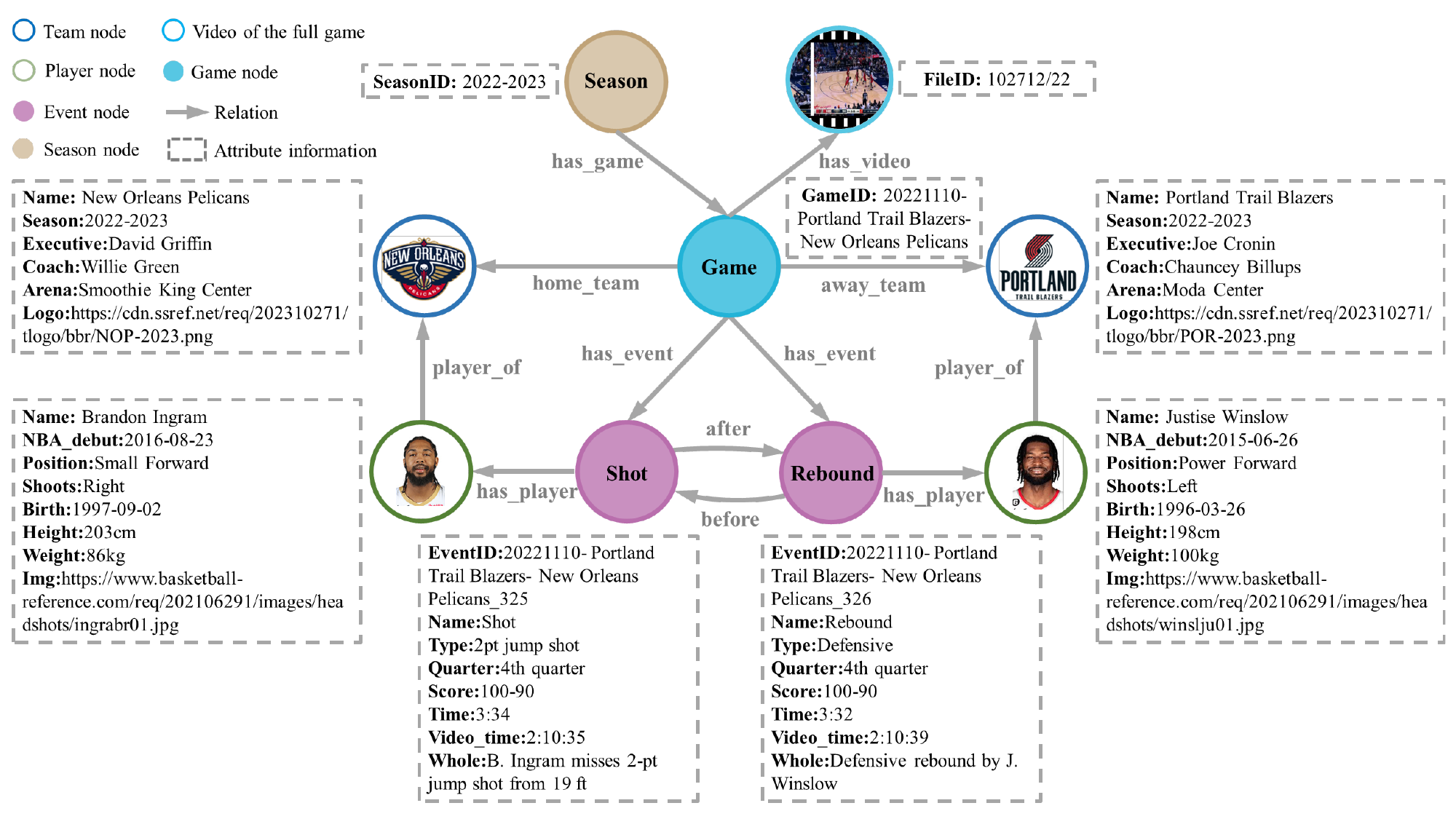}
	\caption{An example of a Multimodal Basketball Game Knowledge Graph.}
	\label{fig:Figure 2}
\end{figure*}

In this work, we focus on the task of knowledge guided entity-aware basketball video captioning (KEBVC), which requires the model to comprehend the content of the provided video and generate text descriptions that include specific entity names and fine-grained actions based on the additional knowledge. We propose a new multimodal knowledge graph supported video captioning benchmark for basketball live text broadcast. This benchmark requires the model to comprehend the video's content and generate event descriptions that incorporate specific entities with additional knowledge. Specifically, we collect 25 NBA full-games multimodal data from professional basketball platforms in 2022-2023 season, including events, players' information, teams' information, and videos. Based on the collected data, we construct a multimodal basketball game knowledge graph named KG\_NBA\_2022 to provide the knowledge beyond videos, as depicted in Fig. 2. Subsequently, a multimodal basketball game video captioning dataset named VC\_NBA\_2022 is constructed based on nodes in KG\_NBA\_2022 and relationships among the selected nodes. VC\_NBA\_2022 dataset comprises 9 types of basketball shooting events and 286 players' knowledge (i.e., images and names), with data samples illustrated in Fig. 3.

To effectively utilize knowledge in the VC\_NBA\_2022 dataset, we propose a knowledge guided entity-aware network (KEANet) based on a candidate player list in encoder-decoder form for basketball live text broadcast. KEANet is comprised of 3 separate unimodal encoders for videos, players' images, and players' names, as well as a pre-trained language model that serves as the text decoder for generating descriptions of the given videos. Moreover, the Bi-GRU module is introduced to encode the temporal contextual information, while the entity-aware module is designed to model the associations among the candidate players and highlight the key players. 

The main contributions of this paper are as follows: 
\begin{itemize}
    \item  We provide an in-depth analysis to discover characteristics of basketball domain and construct a multimodal Basketball Game Knowledge Graph. This knowledge graph introduce additional knowledge that extends beyond the video contents. Based on this Knowledge Graph, nodes and relationships are extracted to construct our Multimodal Basketball Game Video Captioning benchmark.
	\item  We develop a knowledge guided entity-aware video captioning network based on a candidate player list in encoder-decoder form, which attentively incorporates key information from additional knowledge to generate text descriptions with specific entity names.
	\item  To validate the generalization of the proposed model, we conduct experiments on the proposed basketball dataset and the football domain dataset Goal \cite{qi2023goal}. The proposed model outperforms existing advanced models, achieving leading performance.
\end{itemize}
\begin{figure*}[htp]
	\centering
	\includegraphics[width=18.0cm,height=5.0cm]{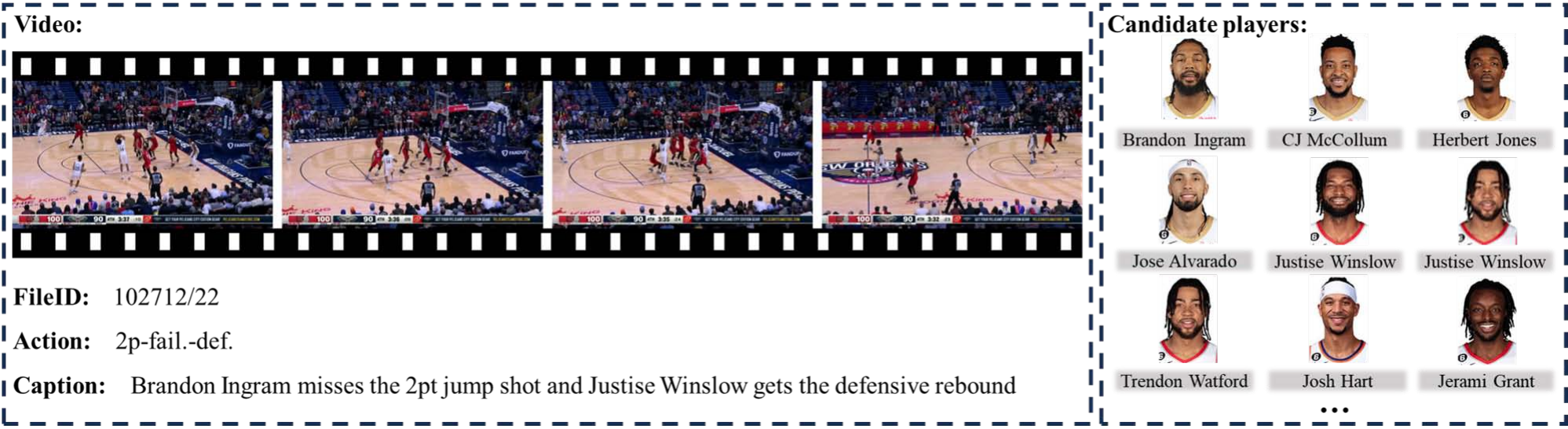}
	\caption{Data sample from the proposed dataset. Each video is annotated by fileid, action type, caption, player images and player names. Each of the players involved in caption as well as their teammates serve as candidate players.}
	\label{fig:Figure 3}
\end{figure*}

\section{Related Works}
\subsection{Computer Vision with Knowledge Graph}
A knowledge graph (KG) is essentially a large-scale semantic network that contains entities, concepts as nodes and various semantic relationships among them as edges \cite{zhu2022multi}. The application of knowledge graphs have greatly promoted the rapid development of computer vision. Zhuo et al. \cite{zhuo2022zero} utilize the mined category-attribute relationships in knowledge graph and the similarity between the seen categories and the unseen ones for more reliable knowledge transfer. One of the main challenges in video-text retrieval task is to identify fine-grained semantic associations between video and text. To address this issue, under the guidance of additional knowledge, Fang et al. \cite{fang2022concept} utilize the associations between concepts to extend new concepts and enrich the representation of videos. This approach enables a more accurate matching of video and text, leading to improved retrieval results. Gu et al. \cite{gu2023text} propose a text with knowledge graph augmented transformer, which integrates additional knowledge and leverages multimodal information to mitigate the challenges posed by long-tail words. 

However, unlike the aforementioned works that extract knowledge from unimodal knowledge graphs, our approach extracts knowledge from a multimodal knowledge graph, utilizing this multimodal knowledge to assist the model in generating descriptions that include specific entity names and fine-grained actions.

\subsection{Video Captioning}
Video captioning is a crucial task in video understanding, where models generate text descriptions for given videos. Some works \cite{li2021value, Bidirectional, luo2020univl, zhang2021open,liu2018sibnet,Emotion,Memory,Contextual} utilize a visual encoder to extract video representations from a set of video frames, and a language decoder transfer these representations to the corresponding descriptions. Recently, CLIP (Contrastive Language-Image Pretraining) \cite{radford2021learning} has demonstrated its superior performance on various visual-linguistic tasks relying on large-scale contrastive pre-training with image-text pairs. Therefore, Clip4Caption \cite{tang2021clip4caption} employs CLIP to acquire the aligned visual-text representation, leading to significant improvements in video captioning performance. Benefiting from the flexibility of the transformer \cite{vaswani2017attention} architecture, SwinBert \cite{lin2022swinbert} introduces Video Swin Transformer \cite{liu2022video} as the video encoder to encode spatial-temporal representations from video frames. None of the previously mentioned works can address the challenge of generating action entity names effectively. In stark contrast, we propose to incorporate additional knowledge based on an encoder-decoder model structure to generate text descriptions with specific entity names and fine-grained actions for basketball live text broadcasts, demonstrating the practicality of our entity-aware video captioning in real applications.

\subsection{Video Captioning Benchmarks}
Existing widely used benchmarks to support the video captioning task include MSVD \cite{chen2011collecting},  MSR-VTT \cite{xu2016msr}, YouTube Hightlight \cite{sun2014ranking} and VATEX \cite{wang2019vatex}. These benchmarks are open-domain databases that enhance the generalization of models. However, these benchmarks suffer from common limitations: the text descriptions are too concise, ignoring the names of specific entities and fine-grained action types. Some works also attempt to enrich benchmarks and develop models to generate text descriptions with more fine-grained information. Fang et al. \cite{fang2020video2commonsense} construct a benchmark annotated with captions and commonsense descriptions. This commonsense benchmark develops models to generate captions as well as 3 types of commonsense descriptions (intention, effect, and attribute). Yu et al. \cite{yu2018fine} propose a fine-grained sports narrative benchmark that focuses more on the detailed actions of the subjects, but this benchmark ignores the names of specific subjects. To assist visually impaired individuals in enjoying movies, Yue et al. \cite{yue2023movie101} construct a large-scale Chinese movie benchmark, which requires models to generate role-aware narration paragraphs when there are no actors speaking. Byeong et al. \cite{Kim2020AutomaticBC} propose a benchmark for automatically generating commentary on baseball games. The descriptions in this benchmark focus on player categories rather than specific player names. To achieve meaningful news summarization, Ayyubi et al. \cite{ayyubi2023video} propose the task of summarizing news video directly to entity-aware captions. They also release a large-scale dataset, to support research on this task. Mkhallati et al. \cite{mkhallati2023soccernet} release a benchmark for soccer broadcasts commentaries. Although the text caption in the benchmark contains specific player names, they do not provide any experimental support for generating a caption with specific names. On the contrary, each specific name is replaced by a specific token (e.g., [TEAM], [COACH], [REFEREE], and [PLAYER]).
\begin{figure*}[ht]
	\centering
	\includegraphics[width=18.0cm,height=8.5cm]{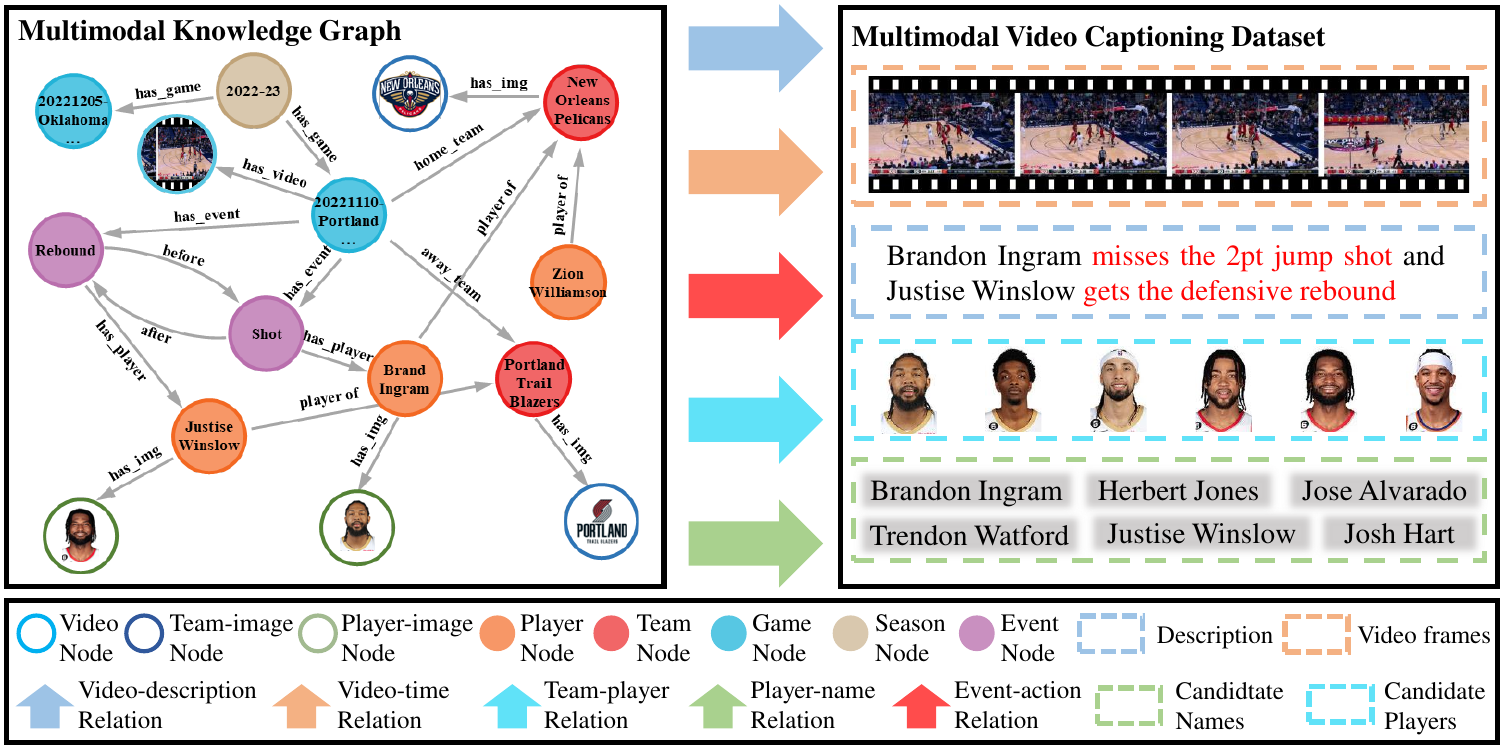}
	\caption{Illustration of extracting relevant data using relationship extraction from the knowledge graph and constructing the dataset.}
	\label{fig:Figure 4}
\end{figure*}

Qi et al. \cite{qi2023goal} utilized a unimodal knowledge graph for real-time soccer commentary, incorporating all pre-match and player-team information without a selection mechanism. This indiscriminate inclusion could introduce noise, affecting caption accuracy.
Contrastingly, our approach first employs a multimodal knowledge graph, integrating both text and visual information, such as player names and images. This enriches the context for video captioning, improving the match with video content. Second, we propose a entity-aware module, modeling associations among candidate players and incorporating key players' knowledge. This effectively reduces noise and enhances caption precision. These advancements distinguish our work from Qi et al.'s \cite{qi2023goal} and other existing models, demonstrating the potential of multimodal knowledge and entity-aware in video captioning tasks.

\section{Proposed Benchmark}
In this section, we first construct a multimodal knowledge graph as the knowledge base, and then extract relevant data and knowledge to construct a multimodal basketball dataset.

\subsection{Multimodal Basketball Game Knowledge Graph}
Our knowledge graph is based on basketball game videos with corresponding event descriptions. To begin the construction of the Multimodal Basketball Game Knowledge Graph, we collect full-game play-by-play data from 50 games, including descriptions of events, the time on the scoreboard corresponding to the events, score records, players' information, and teams' information, from the professional basketball data platform\footnote{https://www.basketball-reference.com}.  The corresponding videos are collected from the basketball broadcast platform\footnote{https://fishkernba.com}. After filtering out videos with low resolution and chaotic content, only 25 games are retained.

To structure the collected data, the following steps are needed to process it: (1) Categorize basketball events. (2) Parse and structure the descriptions of each type of event. (3) Match these event descriptions with the corresponding video timestamps.

By analyzing the play-by-play data, game events can be divided into 9 categories, including ``Foul", ``Rebound", ``Violation", ``Timeout", ``Freethrow", ``Enter game", ``Turnover", ``Jump ball", and ``Shot". Each text description has its owns keyword, such as the description of the ``Foul" event ``Personal foul by G. Temple (drawn by A. Drummond)", where the keyword is ``foul". Therefore, the type of this event is ``Foul". For this semi-structured data, the sentence is parsed into multi-tuples using the character index of the keyword and specific words that appear in the sentence, such as ``drawn by". To incorporate video information into the multi-tuples, the video timestamp is intended to be associated with the text description. OCR is used to identify the time on the scoreboard in each frame and record the timestamp of the current frame. Specifically, we employ the open-source OCR toolkit PaddleOCR\footnote{https://github.com/PaddlePaddle/PaddleOCR} and Tesseract-OCR \cite{smith2007overview} to recognize the time simultaneously. By using multiple OCR toolkits, we can improve the overall accuracy of recognition. This is because one toolkit may succeed where the other fails. In addition to event information, player and team information are also stored in multi-tuples. Through these efforts, a Multimodal Basketball Game Knowledge Graph containing 11489 events and 42870 relationships is constructed. Fig .2 shows an example of the Multimodal Basketball Game Knowledge Graph (MbgKG).
\begin{table*}[tb!]
	\caption{Statistics of the events in KG\_NBA\_2022. ``num.” is the abbreviation of ``number”. Numbers in bold indicate the highest two values.\label{tab:table1}}
	\centering
	\begin{tabular}{cccccccccc}
		\toprule
		Events & Foul & Rebound & Violation & Timeout & Freethrow & Entergame & Turnover & Jumpball & Shot \\ \midrule
		num.   & 986  & \textbf{2649}    & 68        & 271     & 1103      & 1189      & 774      & 49       & \textbf{4400} \\ \bottomrule
	\end{tabular}
\end{table*}

\subsection{Multimodal Basketball Video Game Captioning Dataset}

To construct a dataset that contains videos, text descriptions, and additional knowledge (i.e., player images and names), the pertinent data is extracted from KG\_NBA\_2022 by utilizing the relations within KG\_NBA\_2022. In basketball, shot and rebound events are the most prevalent. Meanwhile, we count the number of different types of events in KG\_NBA\_2022. Table \uppercase\expandafter{\romannumeral1} shows that there are 4400 shot events and 2649 rebound events, which are the most common types of events. Based on the aforementioned common knowledge and statistical data, videos and event descriptions about shot and rebound are extracted from KG\_NBA\_2022 to construct the dataset. To be more realistic, the text descriptions need to be further modified.
\begin{figure}[tb!]
	\centering
	\includegraphics[width=8.8cm,height=3.9cm]{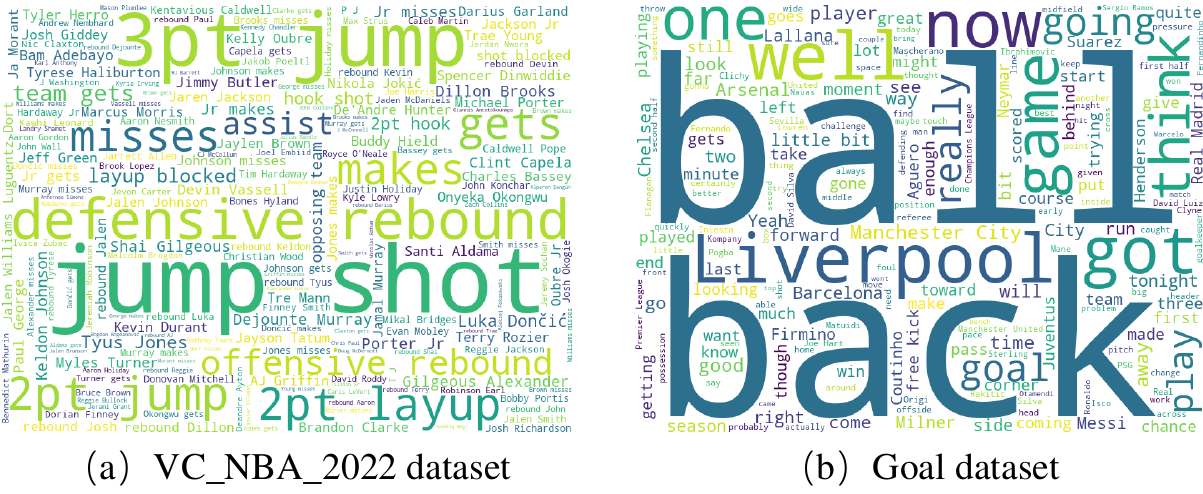}
	\caption{Word cloud of VC\_NBA\_2022 and Goal  datasets. The bigger
		the font, the more percentage it occupies.}
	\label{fig:Figure 5}
\end{figure}

When a shot fails to score in basketball, it is often followed by a rebound event. In line with this pattern, we have combined the shot and rebound into a single event. Shots in basketball can be categorized as two-point (2pt) shots, three-point (3pt) shots, and layups. Rebounds are further divided into defensive (def.) rebounds and offensive (off.) rebounds. The classification scheme of our dataset is based on the NBA dataset \cite{yan2020social} used for group activity recognition. However, unlike the NBA dataset, where all videos are 6 seconds long and divided into 72 frames, we consider that different events may have different durations. Consequently, in our dataset, videos of different lengths are uniformly divided into 72 frames. Additionally, we have incorporated text descriptions and player information into our dataset.

As shown in Fig. 4, the event-action relation in the graph is used to extract 9 kinds of events, and the video-time relation in the graph is used to obtain the video timestamp of the event. Specifically, we first roughly estimate the start and end timestamps of the event based on the existing timestamp. Each clip undergoes a manual review process to ensure accurate start and end timestamps. Subsequently, the video-description relation is used to obtain the text description, which is then matched with the corresponding video clip. We extract player-related information (player images and player names) from KG\_NBA\_2022 through the team-player and player-name relations. Since each event involves certain players, only the involved individuals and their teammates need to be considered as candidate players for video annotation. 

Using the sentence ``B. Ingram misses 2-pt jump shot from 19 ft and defensive rebound by J. Winslow" as an example, we will explain our text modification process. Due to the difficulty in generating distance in video captioning task, the ``from 19 ft" in the sentence is removed. Based on the attribute information of the player in KG\_NBA\_2022, the abbreviated name (``J. Winslow") in the sentence is replaced with the full name (``Justise Winslow"). This change might increase the complexity of name generation but is more aligned with real-world scenarios. To improve text fluency, alterations are made based on the character index. Specifically, we identify the index of keywords like "defensive" and apply a rule (``defensive/offensive rebound by SOMEONE" $\Longrightarrow$ ``SOMEONE gets the defensive/offensive rebound") to revise the sentence. Finally, there are a total 3977 videos\footnote{The scale of the current dataset is not large enough, which is emergent to be enriched in our subsequent research.}, each of which belongs to one of 9 kinds of events. Each video has one text description and several candidate players information (images and names). We randomly select 3162 clips for training and 786 clips for testing. We also provide word-cloud-based statistics in Fig. 5 (a) to reveal the relative amount of different words. It shows that the top-4 subjects in VC\_NBA\_2022 are ``jump”, ``shot”, ``3pt”, and ``defensive”, followed by ``rebound”, ``2pt”, ``layup”, and ``makes”. Table \uppercase\expandafter{\romannumeral2} shows the sample distributions across different labels of events.

In Table \uppercase\expandafter{\romannumeral3}, the comparison of our dataset with MSR-VTT, YouCook and ActivityNet Captions further demonstrates the fine-grained details of our captioning annotations. VC\_NBA\_2022 has the most sentences per second of 0.182, while the other datasets are all below 0.1, this reflects that our dataset has more detailed information in descriptions. Moreover, VC\_NBA\_2022 has 1.73 verbs in a sentence on average, higher than 1.41 for ActivityNet Captioning and 1.37 for MSR-VTT. Similarly, the verb ratio of VC\_NBA\_2022 which is computed by dividing the total number of verbs by the total number of words in the sentence is also much higher than other 3 datasets. This highlights that our dataset is primarily focused on the fine-grained actions of the subjects, aligning with our original intent.
\begin{figure*}[t!]
	\centering
	\includegraphics[width=18.1cm,height=7.1cm]{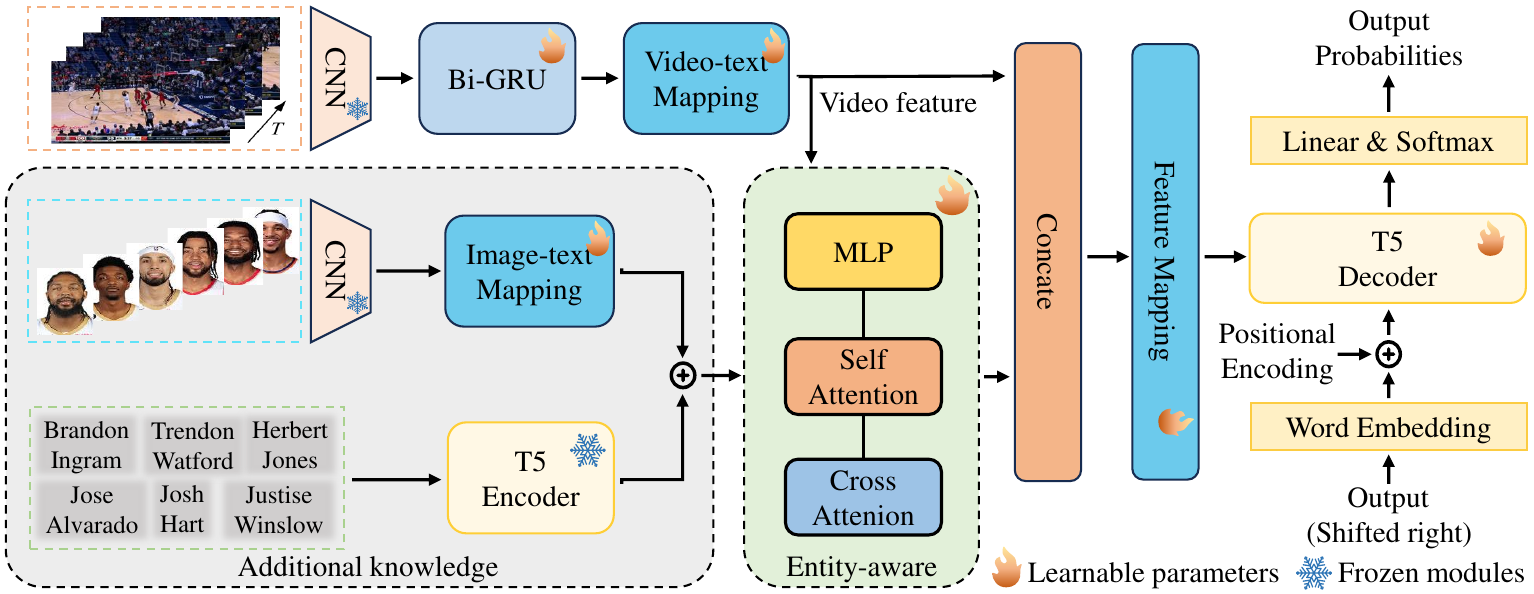}
	\caption{The architecture of knowledge guided entity-aware video captioning network (KEANet) based on a candidate player list in encoder-decoder form.}
	\label{fig:Figure 6}
\end{figure*}

\begin{table*}[tb!]
	\caption{Statistics of the labels in VC\_NBA\_2022. ``succ." and ``fail." are abbreviations of ``success" and ``failure", respectively.\label{tab:table2}}
	\centering
	\begin{tabular}{cccccccccc}
		\toprule
		Labels     & \begin{tabular}[c]{@{}c@{}}2p\\ -succ.\end{tabular} & \begin{tabular}[c]{@{}c@{}}2p\\ -fail.-off\end{tabular} & \begin{tabular}[c]{@{}c@{}}2p\\ -fail.-def.\end{tabular} & \begin{tabular}[c]{@{}c@{}}2p\\ -layup\\ -succ.\end{tabular} & \begin{tabular}[c]{@{}c@{}}2p\\ -layup\\ -fail.-off.\end{tabular} & \begin{tabular}[c]{@{}c@{}}2p\\ -layup\\ -fail.-def.\end{tabular} & \begin{tabular}[c]{@{}c@{}}3p\\ -succ.\end{tabular} & \begin{tabular}[c]{@{}c@{}}3p\\ -fail.-off.\end{tabular} & \begin{tabular}[c]{@{}c@{}}3p\\ -fail.-def.\end{tabular} \\ \midrule
		Train num. & 469                                                 & 146                                                       & 397                                                        & 442                                                          & 133                                                                  & 251                                                                  & 470                                                 & 202                                                        & 652                                                         \\
		Test num.  & 95                                                  & 41                                                        & 95                                                         & 108                                                          & 32                                                                   & 67                                                                   & 125                                                 & 61                                                         & 162                                                         \\ \bottomrule
	\end{tabular}
\end{table*}
\begin{table}[tb!]
	\caption{Comparisons of different video caption datasets. Numbers in 
		bold indicate the highest value.\label{tab:table3}}
	\centering
	\begin{tabular}{cccc}
		\toprule
		Dataset                                                        & \begin{tabular}[c]{@{}c@{}}Sentences\\ per second\end{tabular} & \begin{tabular}[c]{@{}c@{}}Verbs per\\ sentence\end{tabular} & \begin{tabular}[c]{@{}c@{}}Verb\\ ratio\end{tabular} \\ \midrule
		MSR-VTT                                                        & 0.067                                                          & 1.37                                                         & 14.8\%                                               \\
		YouCook                                                        & 0.056                                                          & 1.33                                                         & 12.5\%                                               \\
		\begin{tabular}[c]{@{}c@{}}ActivityNet\\ Captions\end{tabular} & 0.028                                                          & 1.41                                                         & 10.4\%                                               \\ 
		\textbf{VC\_NBA\_2022}                                                           & \textbf{0.182}                                                         & \textbf{1.73}                                                         & \textbf{14.8\%}                                               \\ \bottomrule
	\end{tabular}
	
\end{table}
\section{Knowledge-based Entity-aware Basketball Video Captioning}

\subsection{Problem Formulation}
To generate precise and concise text descriptions of the basketball live text broadcast, we introduce the concept of KEBVC task. In this task, the model is required to comprehend the content of the provided video and generate text descriptions that include specific entity names and fine-grained actions based on the additional knowledge. This task can be formulated as: given the basketball video $V_{b}$, the objective is to select video-related player knowledge $K_{p}=s\left ( V_{b} \right )$ and generate the text description $D_{v,k}$.
\begin{equation}
{\rm KEBVC}:D_{v,k} =m\left ( V_{b},K_{p}  \right  ) =m\left ( V_{b},s\left ( V_{b} \right )  \right ),
\end{equation}
where $s\left ( \cdot  \right )$ and $m\left ( \cdot  \right )$ denote the models' abilities on aligning video and knowledge (player information) and transferring video and knowledge to the text description, respectively.

\begin{table*}[tb!]
	\caption{Combined inference performance on VC\_NBA\_2022. Numbers in 
		bold indicate the best performance.\label{tab:table4}}
	\centering
	\begin{tabular}{ccccccccc}
		\toprule
		Model       & CIDEr & METEOR & Rouge-L & BLEU-1 & BLEU-2 & BLEU-3 & BLEU-4 & RoleF1  \\ \midrule
		V2C          & 13.7  & 18.7   & 45.9    & 50.1   & 39.5   & 27.0   & 14.9   & 0.0    \\
		Clip4Caption & 70.4  & 26.7   & 51.2    & 49.1   & 42.5   & 35.4   & 28.8   & 8.3    \\
		SwinBert     & 69.1  & 26.5   & 49.0    & 47.8   & 41.4   & 34.5   & 28.4   & 7.4    \\ 
		\textbf{KEANet}          & \textbf{138.5} & \textbf{28.0}   & \textbf{54.9}    & \textbf{53.1}   & \textbf{46.4}   & \textbf{38.8}   & \textbf{32.4}   & \textbf{20.6} \\ \bottomrule
	\end{tabular}
\end{table*}

\begin{figure}[tb!]
	\centering
	\includegraphics[width=8.5cm,height=4.3cm]{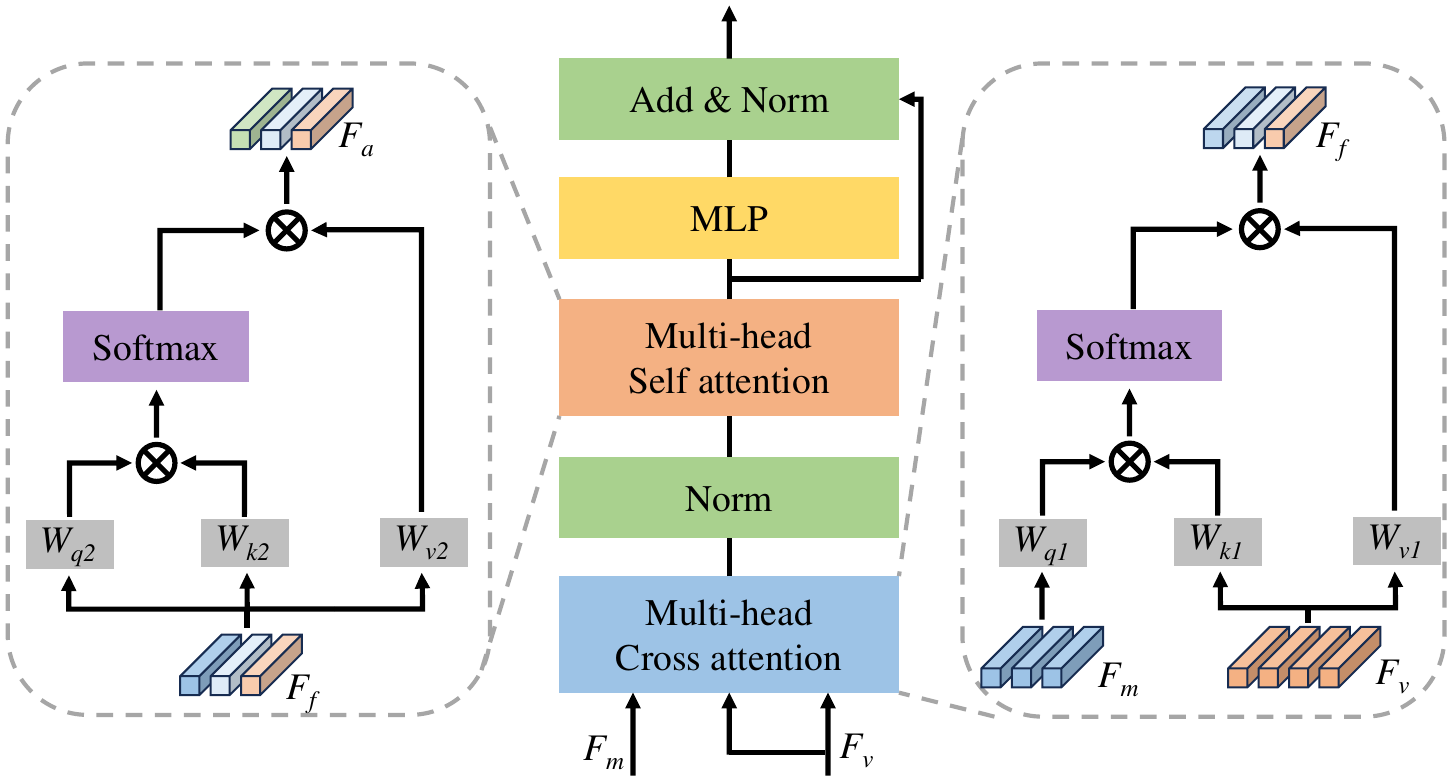}
	\caption{Detailed architecture of the entity-aware module.}
	\label{fig:Figure 7}
\end{figure}
\subsection{Proposed Model}
For the ${\rm KEBVC}$ task, we propose a knowledge guided entity-aware video captioning network (KEANet) based on a candidate player list in encoder-decoder form. The overall structure of KEANet framework is shown in Fig. 6. Given the raw video frames which are of size $T\times 3\times H\times W$, consisting of $T$ frames and each has $3\times H\times W$ pixels, we feed them into the CNN-based visual encoder in KEANet and extract the video feature $F_{T} \in {\rm \mathbb{R}}^{T\times D_{r} \times h\times w}$. $D_{r}$ is the hidden size of visual feature. The global representations of $F_{T}$ are then fed into Bi-GRU module to further encode temporal contextual information and obtain the feature $F_{v}\in {\rm \mathbb{R}}^{T\times D}$.
\begin{equation}
F_{v}=W_{1} \left ( {{\rm GRU} \left ( {\rm AvgPool}\left ( F_{T}  \right )  \right ) } \right ),
\end{equation}
where $W_{1} \in {\rm \mathbb{R}}^{D_{r} \times D}$ is the linear mapping layer, which maps the video feature to text space. $D$ is the hidden size of the decoder module. ${\rm GRU}\left ( \cdot  \right )$ denotes the Bi-GRU module and ${\rm AvgPool}\left ( \cdot  \right )$ denotes the average pooling layer. 

Each video has the corresponding candidate $N$ players' information as the additional knowledge to assist the generation of text with the specific entity names. Each player image is of size $3\times H_{p}\times W_{p}$. For candidate players' images, we use the CNN-based visual encoder to extract the features $F_{N} \in {\rm \mathbb{R}}^{N\times D_{r} \times h_{p}\times w_{p}}$. The global features  $F_{p} \in {\rm \mathbb{R}}^{N\times D}$ of $N$ images are obtained by (\ref{pp}).
\begin{equation}
\label{pp}
F_{p}=W_{2} \left ( {\rm AvgPool}\left ( F_{N}  \right )  \right )),
\end{equation}
where $W_{2} \in { \mathbb{R}}^{D_{r} \times D}$ is the linear mapping layer, which maps the image feature to text space.

\begin{table}[tb!]
	\caption{Combined inference performance on Goal. Numbers in 
		bold indicate the best performance.\label{tab:table5}}
	\centering
	\begin{tabular}{ccccc}
		\toprule
		Method       & CIDEr & METEOR & Rouge-L & BLEU-1 \\ \midrule
		V2C          & 0.13  & 2.11   & 3.42    & 3.38   \\
		Clip4Caption & 2.17  & 5.01   & 5.45    & 5.71   \\
		SwinBERT     & 2.20  & 5.12   & 5.32    & 5.65   \\
		\textbf{KEANet}          & \textbf{3.68}  & \textbf{6.38}   & \textbf{10.45}   & \textbf{14.88}  \\ \bottomrule
	\end{tabular}
\end{table}
The text encoder of large language model T5~\cite{raffel2020exploring} is used to transform candidate names into a sequence of embeddings $T_{n} \in  {\mathbb{R}}^{N \times D}$. After that, the image features $F_{p}$ and name features $T_{n}$ are added by corresponding positions to obtain multimodal player features $F_{m}\in  {\mathbb{R}}^{N \times D}$. Basketball events involve interactions between multiple players. For example, in ``Myles Turner makes the 2pt layup with an assist from Tyrese Haliburton", ``Myles Turner" scores a two-point layup with the assist of ``Tyrese Haliburton". Therefore, we design the entity-aware module to highlight key players from a candidate player list. The architecture of entity-aware module is shown in Fig. 7. entity-aware module first uses the cross-attention to fuse player features and video features to obtain $F_{f}$. This connects the players to the video content, putting the players in a specific scene. $F_{a}$ is sent to the multi-head self-attention sub-module to model the relationship among candidate players. The player features based on attention and the video feature are then concatenated and fed into the decoder ${\rm\Psi_{T5}}\left ( \cdot  \right )$ of T5 to obtain the text description $C$.
\begin{equation}
C={\rm \Psi_{T5}} \left ( W_{3} \left ( {\rm Concat}\left [ {\rm E_{a}} \left ( F_{m}, F_{v}  \right ) ,F_{v}  \right ]  \right )  \right ), 
\end{equation}
where $W_{3} \in {\mathbb{R}}^{D \times D}$ is the linear mapping layer, which maps the concatenated feature to the vector space of the T5 model. ${\rm Concat}\left [ ,  \right ]$ denotes the concatenation function in Python, and ${\rm E_{a}\left ( \cdot  \right ) }$ denotes the entity-aware module.



The Bi-GRU module, 3 linear layers, entity-aware module and text decoder are trained to maximize the log-likelihood over the training set given by (\ref{loss}).
\begin{equation}
\label{loss}
\mathcal{L}_{\Theta } =\sum_{t=1}^{N_{C} } log{\rm P_{r} }\left ( y_{t} |y_{t-1} ,\left [ F_{m} ,F_{v}  \right ] ;\Theta  \right ),
\end{equation}
where $y_{t}$ denotes the one-hot vector probability of each word at time $t$. $N_{C}$ denotes the length of the caption. And $\Theta$ denotes the learnable parameters.

\section{Experiments}
To verify the performance of the proposed model, KEANet is compared with 
the advanced video captioning models on VC\_NBA\_2022 and Goal \cite{qi2023goal}. We further conduct the ablation experiments to verify the effectiveness of each conponent in KEANet.

\subsection{Implementation Details}
The proposed KEANet uses ResNet-18~\cite{he2016deep} pretrained on the ImageNet dataset \cite{russakovsky2015imagenet} as the visual encoder, which hidden size $D_{r}$ is 512. KEANet is trained on the proposed VC\_NBA\_2022 and Goal datasets with 100 training epochs. In these datasets, $T$ frames are sampled by using segment-based method \cite{wang2016temporal}. Each frame size is $1280 \times 720$. A high resolution of video frame is helpful for the model to understand the players' action in the video. Concretely, $T$ is set to 18. The size of player's image is $180 \times 120$ and the number of candidate players in each video is not fixed. For the large language model T5, the hidden size is 768. It is worth noting that the parameters of the visual encoder and text encoder are frozen during training. During the training stage, the model is optimized by ADAM \cite{kingma2014adam} with the learning rate of 3e-5 and weight decay of 1e-4. Beam search with a beam size of 2 is used for inference. The proposed KEANet is implemented with Python 3.9 and PyTorch 1.12, and is performed on a server with an Nvidia 3090ti GPU.
\begin{figure*}[tb!]
	\centering
	\includegraphics[width=17.5cm,height=19.0cm]{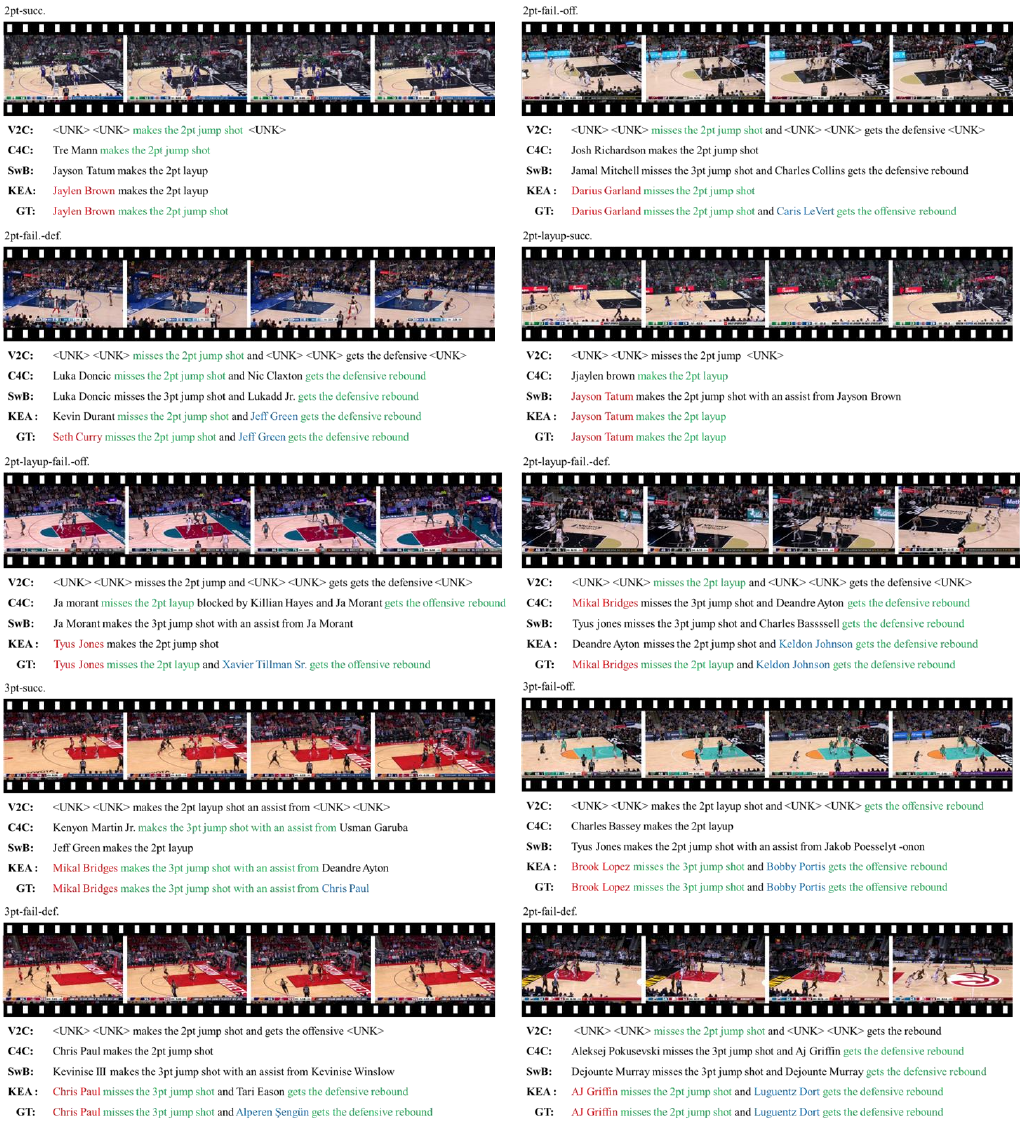}
	\caption{Qualitative results on VC\_NBA\_2022 dataset. (V2C: Video2Commonsense; C4C: Clip4Caption; SwB: SwinBert; KEA: our proposed model; GT: the ground truth). The different specific entity names are marked in {\color{red}red} and {\color{blue}blue}, respectively. And the fine-grained actions are marked {\color{green}green}. Since V2C does not have its own tokenizer and vocabulary list, it cannot decode names. So names are replaced with the special token $<$UNK$>$.}
	\label{fig:Figure 8}
\end{figure*}
\subsection{Evaluation metrics}
Existing video captioning benchmarks mainly adopt ngram-based metrics, including CIDEr \cite{vedantam2015cider}, METEOR \cite{banerjee2005meteor}, Rouge-L \cite{lin2004rouge}, and BLEU \cite{papineni2002bleu}. BLEU evaluates the quality of the generated text by calculating the n-gram overlap between the generated text and the reference text. Rouge-L pays more attention to generating the longest common subsequence between the generated text and the reference text, and evaluates sentence quality in this way. METEOR takes more factors into account, including stem matching, synonym matching, and word order. CIDEr uses TF-IDF \cite{robertson2004understanding} to assign different weights to n-grams of different lengths, then calculates the cosine similarities of n-grams between generated texts and reference texts and averages them to get the final score. These metrics primarily emphasize the consistency of text rather than the accuracy of semantics. In other words, if the generated text differs from the reference in structure or vocabulary, it may be penalized by these n-gram based evaluation metrics, even if the conveyed information is completely correct. However, for entity names, they cannot be replaced with the similar words. To accurately evaluate the performance of specific entity name generation, we introduce RoleF1 score (F1 score of entity name generation) from the work \cite{yue2023movie101}. For RoleF1, we extract entity names from the ground truth and the generated texts. Recall is used to measure how the generated texts covers the entities appearing in the game video clip. Taking into account that these generated entity names could also come from the model's hallucinations, such as from incorrect games, we will also consider Precision. Finally, RoleF1 score is calculated with Precision and Recall.
\begin{figure}[tb!]
	\centering
	\includegraphics[width=7.0cm,height=5.4cm]{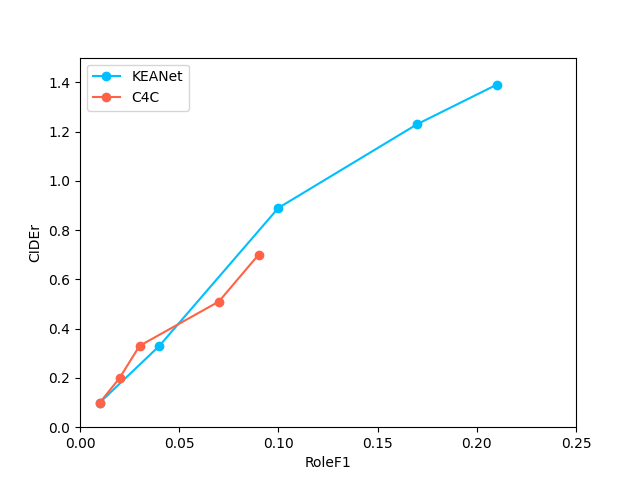}
	\caption{The relationship between metric CIDEr and RoleF1. C4C denotes Clip4Caption.}
	\label{fig:Figure 9}
\end{figure}
\subsection{Results on VC\_NBA\_2022}
KEANet is compared with three advanced video captioning models, including
V2C \cite{fang2020video2commonsense}, Clip4Caption \cite{tang2021clip4caption}, and SwinBert \cite{lin2022swinbert}. V2C is the Transformer-based model that generates relevant commonsense descriptions of the given video.  Clip4Caption employs CLIP to acquire the aligned visual-text representation for better generating the text descriptions. SwinBert introduces Video Swin Transformer to encode spatial-temporal representations from video frames. V2C, Clip4Caption, and SwinBert take only videos as input, and are all trained and tested on our proposed dataset.

As shown in Table \uppercase\expandafter{\romannumeral4}, KEANet outperforms other 3 models by a large margin on CIDEr. This notable performance can be attributed to the way CIDEr calculates cosine similarities of n-grams between the generated text and reference text. Highly accurate name prediction contributes to a higher cosine similarity between the n-grams of the generated text and the reference text. V2C, Clip4Caption, and SwinBert exhibit lower performance in accurately generating names, resulting in much lower CIDEr scores. Although KEANet achieves the best performance across all metrics, the differences are not as pronounced in METEOR, Rouge-L, and BLEU. This can be attributed to the concise nature of the text descriptions in the dataset, primarily consisting of names and fine-grained actions. Notably, KEANet outperforms SwinBert 13.2\% on RoleF1 and outperforms Clip4Caption 12.3\% on RoleF1. The higher RoleF1 indicates better name prediction. These results underscore the capability of our model to generate accurate names and fine-grained actions in live text broadcast task. 

Fig. 8 shows the qualitative results on VC\_NBA\_2022 dataset, including generation results of V2C, Clip4Caption, SwinBert and our KEANet model. V2C, Clip4Caption and SwinBert can correctly generate some actions but fail to generate correct entity names because these names never appear during training. However, with the help of the additional knowledge, our KEANet could well relate entities in video clips with the fine-grained actions. These cases demonstrate that our newly proposed KEANet model more consistently with the requirement of practical application.

Further, we study the relation between the metric CIDEr and RoleF1. To this end, we plot the performance of KEANet and Clip4Caption with different RoleF1 scores in Fig. 9. The plot shows that this relationship is approximately linear: more accurate entity name generation imply better model performance.

\subsection{Results on Goal}
Goal is a benchmark which contains over 8.9k soccer video clips, 22k sentences and 42k knowledge triples. These sentences were converted from the commentator's audios. This type of comment sentences have a certain degree of colloquialism and are relatively long. Therefore, this dataset poses certain challenges. On this basis, we filter out sentences and videos that do not contain entity names. We modify the format of the dataset to be the same as VC\_NBA\_2022. In addition, we save the names of teams and players in the dataset. The revised Goal dataset is shown in Fig. 10.

KEANet is compared with three advanced video captioning models, including V2C, Clip4Caption, and SwinBert.  V2C, Clip4Caption, and SwinBert take only videos as input, and are
all trained and tested on the revised Goal dataset. As shown in Table \uppercase\expandafter{\romannumeral5}, we compare the performance of V2C, Clip4Caption, SwinBERT, and KEANet on several metrics including CIDEr, METEOR, Rouge-L, and BLEU-1. KEANet is supported by additional knowledge that it can generate text descriptions with entity names. Therefore, the KEANet outperforms other models in all metrics. 

Fig. 11 shows the qualitative results on Goal dataset, including generation results of V2C, Clip4Caption, SwinBert and our KEANet. Due to the lack of relevant tokenizer tools in V2C, it cannot process text with entity names and can only use special symbols to replace specific entity names. Compared to the other three models, KEANet can generate correct entity names and partial actions. The above results indicate that, even in challenging tasks such as sports commentary, with the support of additional knowledge, the model can generate text with entity names.

\begin{table*}[tb!]
	\caption{Ablation study on MbgVC. Ki and Kn denote the player images knowledge and player names knowledge, respectively. Ea. is the entity-aware module. Bi-GRU is bi-directional GRU module. Numbers in 
		bold indicate the best performance.\label{tab:table6}}
	\centering
	\begin{tabular}{cccccccccc}
		\toprule
		Model & Ki & Kn & Ea. & Bi-GRU & CIDEr & METEOR & Rouge-L & BLEU-4 & RoleF1  \\ \midrule
		\ding{172}     &    &    &    &                    & 20.1  & 20.6   & 40.9    & 23.0   & 3.4    \\
            \ding{173}     &    &    &    & \checkmark         & 23.5  & 22.7   & 43.1    & 24.4   & 3.6    \\
		\ding{174}     & \checkmark  &    &       &       & 18.0  & 20.0   & 38.0    & 23.2   & 3.3    \\
		\ding{175}     &    & \checkmark  &       &       & 110.6 & 25.2   & 49.1    & 27.3   & 12.4    \\
		\ding{176}     & \checkmark  & \checkmark  &       &       & 115.7 & 25.6   & 53.1    & 28.1   & 14.8    \\
		\ding{177}     & \checkmark  & \checkmark  & \checkmark     &       & 119.3 & 26.8   & 53.3    & 29.5   & 16.5    \\
		\ding{178}     & \checkmark  & \checkmark  &       & \checkmark     & 122.2 & 27.3   & 54.6    & 30.5   & 17.1    \\
		\ding{179}     & \checkmark  & \checkmark  & \checkmark     & \checkmark     & \textbf{138.5} & \textbf{28.0}   & \textbf{54.9}    & \textbf{32.4}   & \textbf{20.6} \\ \bottomrule
	\end{tabular}
\end{table*}
\begin{figure*}[htp]
	\centering
	\includegraphics[width=18.0cm,height=5.0cm]{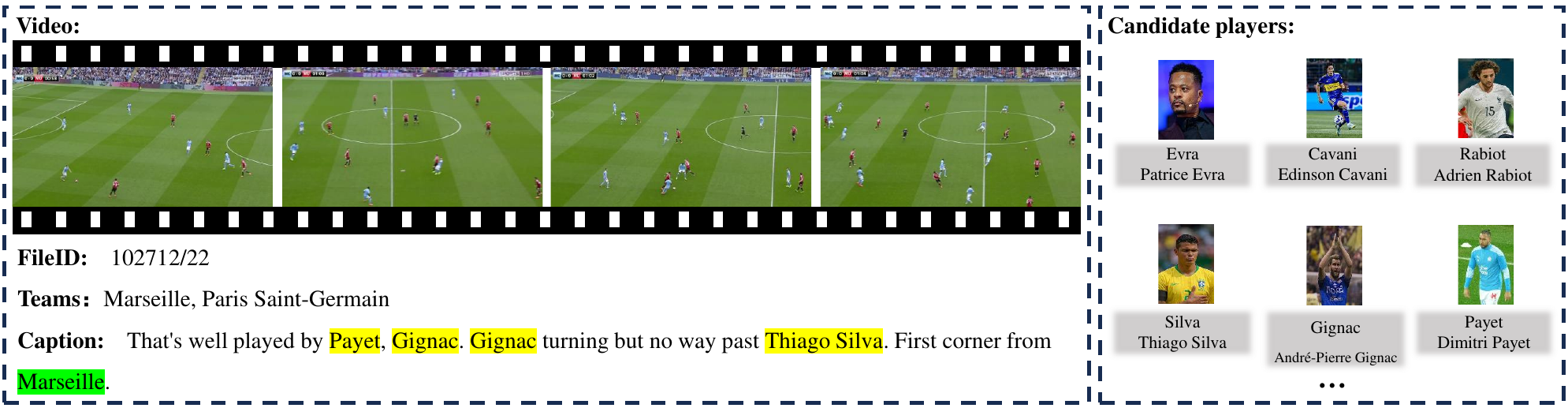}
	\caption{Data sample from the revised Goal dataset. Each video is annotated by fileid, team\_id, caption, player images and player names. Each of the players involved in caption as well as their teammates serve as candidate players. The caption in this sample includes the names of \sethlcolor{yellow}\hl{players} and \sethlcolor{green}\hl{team}.} 
	\label{fig:Figure 10}
\end{figure*}
\begin{figure*}[tb!]
	\centering
	\includegraphics[width=18.0cm,height=7.8cm]{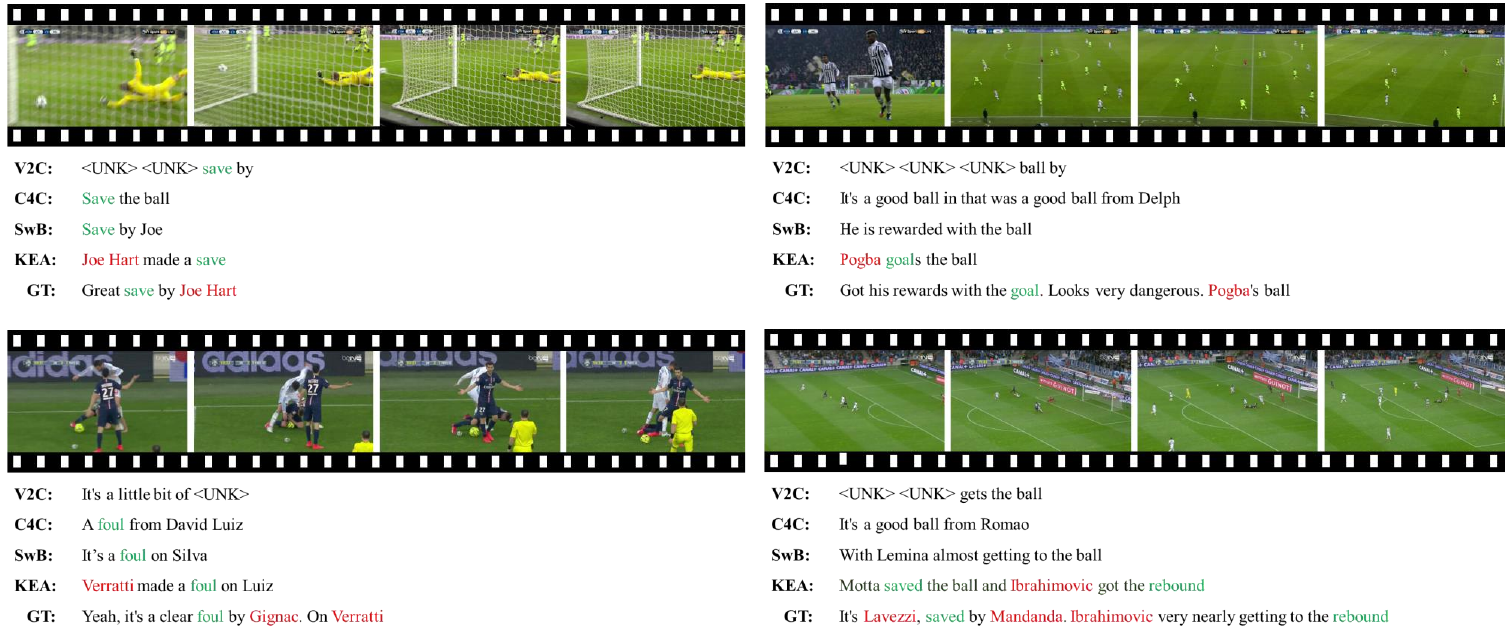}
	\caption{Qualitative results on Goal dataset. (V2C: Video2Commonsense; C4C: Clip4Caption; SwB: SwinBert; KEA: our proposed model; GT: the ground truth). The different specific entity names are marked in {\color{red}red}. And the fine-grained actions are marked {\color{green}green}. Since V2C does not have its own tokenizer and vocabulary list, it cannot decode names. So names are replaced with the special token $<$UNK$>$.} 
	\label{fig:Figure 11}
\end{figure*}
\subsection{Ablation Study}
To verify the contribution of the additional knowledge and other modules in KEANet, we also perform an ablation study by progressively adding these as input. The results of ablation study are shown in Table \uppercase\expandafter{\romannumeral6}. Model \ding{172} consists of Resnet-18 and T5 decoder, with its input being solely video. Model \ding{173} adds the Bi-GRU module on top of Model \ding{172}. The Bi-GRU models the temporal contextual relationships between video frames, enabling the model to better understand the dynamic information in videos and predict more accurate action categories. Therefore, Model \ding{173} has shown improvements in all metrics relative to Model \ding{172}, except for the RoleF1 score. From the comparison results between of model \ding{172} and model \ding{174}, it can be observed that adding players' images to model based on \ding{172} does not improve the model's performance. During the training stage, if the model has not been exposed to enough specific names, or has not learned how to infer names from video and image features, it may not be able to generate these names. The comparison results between model \ding{172} and model \ding{175} highlight a significant improvement by solely adding players’ names to the model based on \ding{172}. Providing a list of names as input allows the model to effectively integrate this information with the video features, leading to more accurate generation of names. This is likely providing the model with additional context information to help it make more accurate predictions. From the comparison results between model \ding{175} and model \ding{176}, we can find that player images can emphasize the roles based on entity names and improve the names' prediction. From the comparison results of models \ding{176}-\ding{179}, it can be observed that adding entity-aware and Bi-GRU modules can improve the performance of the model. The entity-aware module can model the associations among players and focus on key players. The Bi-GRU module can model the temporal contextual information of video frame features, capturing the action information. From the above results, the additional players knowledge brings significant gains in entity awareness. The model can generate text descriptions with specific player names and fine-grained actions through additional knowledge and temporal modeling.

\begin{figure}[tb!]
	\centering
	\includegraphics[width=8.3cm,height=8.0cm]{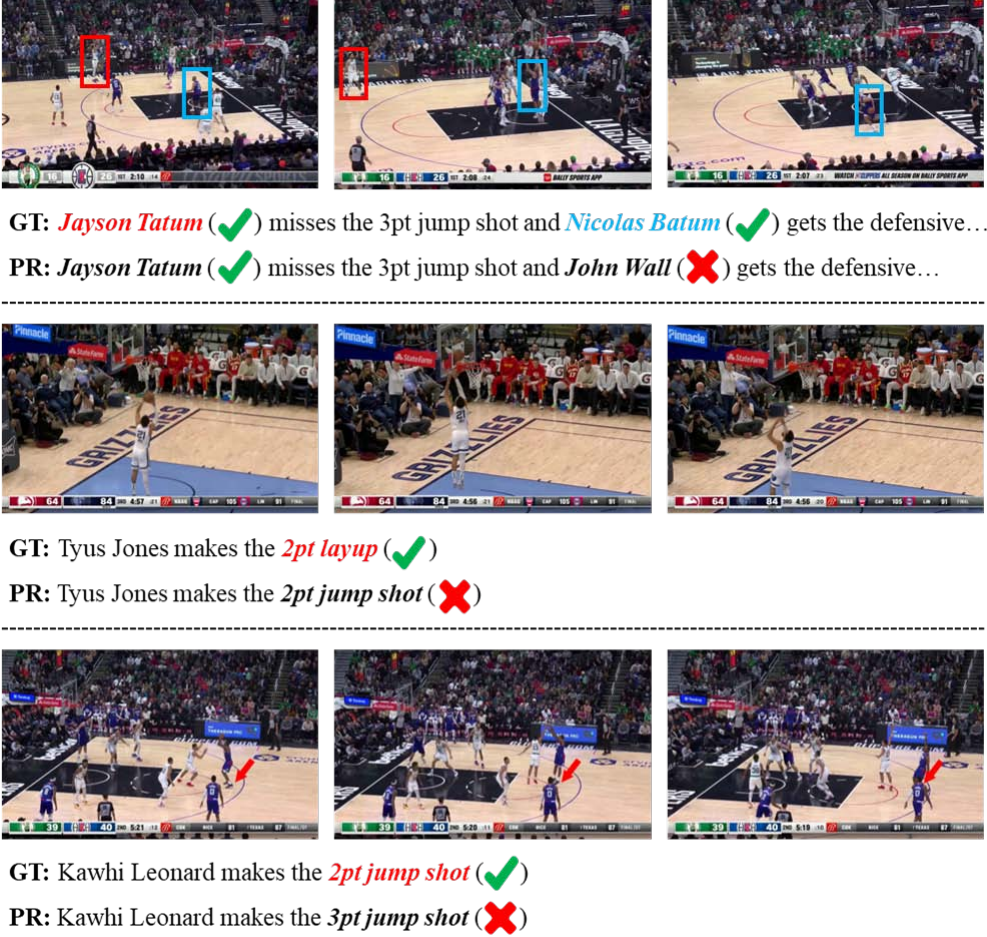}
	\caption{Representative error cases of the generated captions, which correspond to the player mismatching, action confusion and lack distance-aware perception. GT denotes the ground truth description and PR denotes the generated description.}
	\label{fig:Figure 12}
\end{figure}
\subsection{Error Analysis}
The error analysis are conducted by presenting cases of KEANet on proposed VC\_NBA\_2022 dataset. As shown in Fig. 12,  we show several errors. (1) Player Mismatching: when the number of players in the sentence is more than one, the model may decode the name of one of the players incorrectly. (2) Action Confusion: model tends to confuse similar-looking actions, such as layups and close-range two-point jump shots. (3) Lack Distance-aware Perception: model tends to confuse three-point jump shots and long-range two-point jump shots. For example, the player appears to be shooting a three-point shot, but is actually inside the three-point line. 

Given the above result, we discuss some potential ways for developing an advanced models for KEBVC task. First, beyond the current object detection, it is necessary to enhance the model's ability to understand and handle names. Second, regarding action confusion, this may be because the model is not sensitive enough to subtle differences in the video to accurately distinguish similar actions. We need to improve the feature extraction abilities of the model to more accurately recognize similar-looking actions. Third, for the lack of distance perception, this may be because the model has trouble processing spatial information, especially when judging the relative distance of players from the basket. This indicates that we need to further improve the spatial awareness of our model.

\section{Conclusion}
In this paper, we introduce the task of knowledge guided entity-aware basketball video captioning (KEBVC) for live text  broadcast. To achieve this task, we further construct a multimodal basketball game knowledge graph (KG\_NBA\_2022) to provide the additional knowledge. Through partial relationships in the knowledge graph, a new multimodal basketball game video captioning dataset is then constructed. Furthermore, our experiments validate the importance of additional knowledge, including player images and entity names, for entity-aware video captioning.

\bibliographystyle{IEEEtran}  
\bibliography{main}

\end{document}